\documentclass[conference]{IEEEtran}

\usepackage{amsmath,amsfonts,bm}

\def\eqref#1{equation~\ref{#1}}

\def\1{\bm{1}}

\DeclareMathAlphabet{\mathsfit}{\encodingdefault}{\sfdefault}{m}{sl}
\SetMathAlphabet{\mathsfit}{bold}{\encodingdefault}{\sfdefault}{bx}{n}

\newcommand{\sigmoid}{\sigma}

\DeclareMathOperator*{\argmax}{arg\,max}

\usepackage{hyperref}
\usepackage{url}
\usepackage{color}
\usepackage{xspace}
\usepackage{amsmath,amsfonts,amssymb}
\usepackage{booktabs}
\usepackage{subcaption}
\usepackage{graphicx}
\usepackage{makecell}
\usepackage{algorithm}
\usepackage{cite}
\usepackage{multirow}
\usepackage{cleveref}
\usepackage[noend]{algpseudocode}
\makeatletter
\def\BState{\State\hskip-\ALG@thistlm}
\makeatother

\newcommand{\bl}{{\sf Anomaly Link Discovery}\xspace}
\newcommand{\bls}{{\sf ALD}\xspace}
\newcommand{\blb}{{\sf Katz Index}\xspace}
\newcommand{\blbs}{{\sf Katz}\xspace}
\newcommand{\linkp}{{\sf LinkPred}\xspace}
\newcommand{\linkps}{{\sf LP}\xspace}

\newcommand{\graphg}{{\sf GraphGenDetect}\xspace}
\newcommand{\graphgs}{{\sf GGD}\xspace}
\newcommand{\abnorm}{{\sf OutlierDetect}\xspace}
\newcommand{\abnorms}{{\sf OD}\xspace}
\newcommand{\ensemb}{{\sf EDoG}\xspace}
\newcommand{\ensembs}{{\sf EDoG}\xspace}

\newif\ifsubmit
\submitfalse

\ifsubmit
\newcommand{\xiaojun}[1]{}
\newcommand{\bo}[1]{}
\newcommand{\lesong}[1]{}
\newcommand{\yue}[1]{}
\newcommand{\wh}[1]{}
\else
\newcommand{\xiaojun}[1]{{\color{blue}[Xiaojun: #1]}}
\newcommand{\bo}[1]{{\color{red}[Bo: #1]}}
\newcommand{\lesong}[1]{{\color{pink}[Le: #1]}}
\newcommand{\yue}[1]{{\color{green}[Yue: #1]}}
\newcommand{\wh}[1]{{\color{cyan}[Warren: #1]}}

\fi

\begin{document}
\title{EDoG: Adversarial Edge Detection For Graph Neural Networks} 

\author{\IEEEauthorblockN{
Xiaojun Xu$^{1}$ \quad Yue Yu$^{2}$ \quad Hanzhang Wang$^{3}$ \quad Alok Lal$^{3}$ \quad Carl A. Gunter$^{1}$ \quad Bo Li$^{1}$
}
\IEEEauthorblockA{
$^{1}$University of Illinois at Urbana-Champaign
$^{2}$Georgia Institute of Technology
$^{3}$eBay
\\
\texttt{\{\href{mailto:xiaojun3@illinois.edu}{xiaojun3}, \href{mailto:cgunter@illinois.edu}{cgunter}, \href{mailto:lbo@illinois.edu}{lbo}\}@illinois.edu}\\
\texttt{ \href{mailto:yueyu@gatech.edu}{yueyu@gatech.edu} }\\
\texttt{\{\href{mailto:hanzwang@ebay.com}{hanzwang}, \href{mailto:allal@ebay.edu}{allal}\}@ebay.com}\\
}}

\maketitle

\begin{abstract}
Graph  Neural  Networks  (GNNs)  have  been widely applied to different tasks such as bioinformatics, drug design, and social networks. 
However, recent studies have shown that GNNs are vulnerable to adversarial attacks which aim to mislead the node (or subgraph) classification prediction by adding subtle perturbations. In particular, several attacks against GNNs have been proposed by adding/deleting a small amount of edges, which have caused serious security concerns. 
Detecting these attacks is challenging due to the small magnitude of perturbation and the discrete nature of graph data. In this paper, we propose a general adversarial edge detection pipeline \ensembs without requiring knowledge of the attack strategies based on graph generation. Specifically, we propose a novel graph generation approach combined with link prediction to detect suspicious adversarial edges. 
To effectively train the graph generative model, we sample several sub-graphs from the given graph data. We show that since the number of adversarial edges is usually low in practice, with low probability the sampled sub-graphs will contain adversarial edges based on the union bound.
In addition, considering the strong attacks which perturb a large number of edges, we propose a set of novel features to perform outlier detection as the preprocessing for our detection.
Extensive experimental results on three real-world graph datasets including a private transaction rule dataset from a major company and two types of synthetic graphs with controlled properties (e.g., Erdos-Renyi and scale-free graphs) show that \ensembs can achieve above 0.8 AUC against four state-of-the-art unseen attack strategies without requiring any knowledge about the attack type (e.g., degree of the target victim node); and around 0.85 with knowledge of the attack type. \ensembs significantly outperforms traditional malicious edge detection baselines.
We also show that an adaptive attack with full knowledge of our detection pipeline is difficult to bypass it.
Our results shed light on several principles to improve the robustness of GNNs.
\end{abstract}


\section{Introduction}

\begin{figure}
    \centering
    \includegraphics[width=0.9\linewidth]{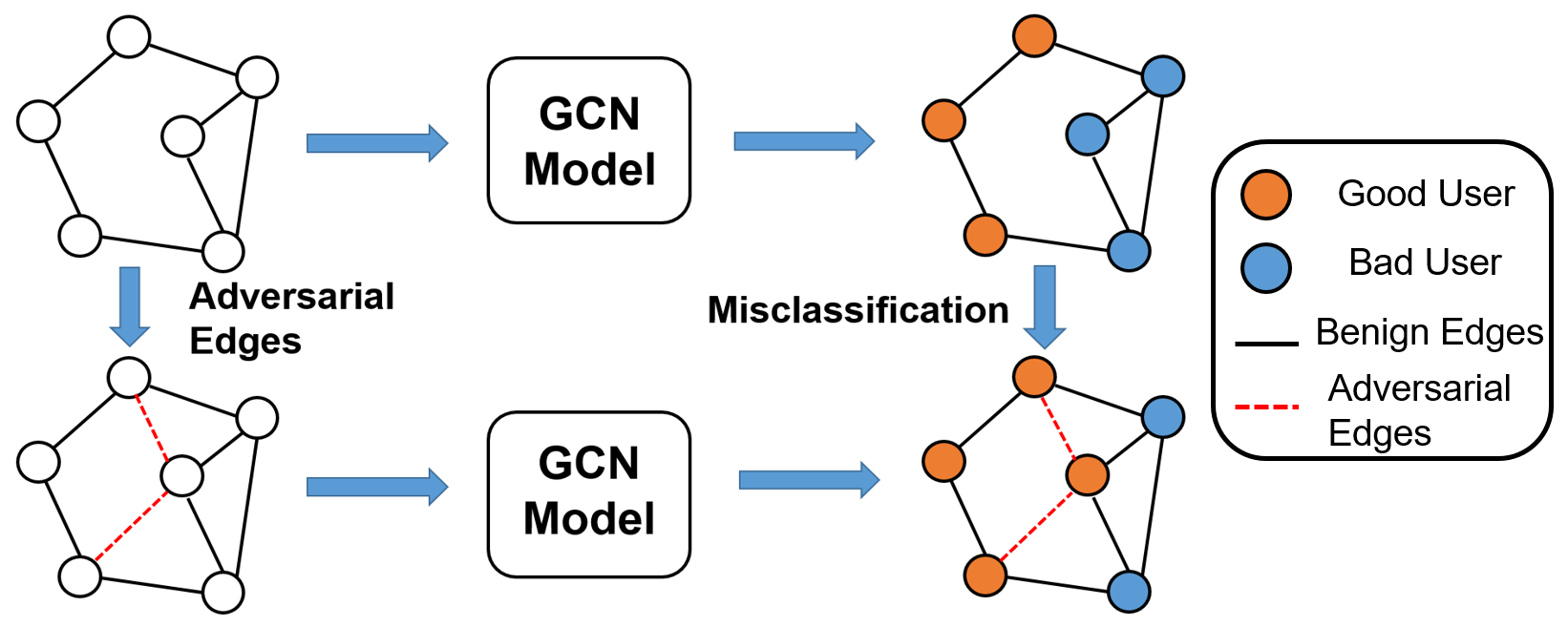}
    \caption{An example of adversarial attack on graph neural networks (GNNs).}
    \label{fig:eg}
    \vspace{-1em}
\end{figure}

Graph neural networks (GNNs) have been widely applied in many real-world tasks, such as drug screening~\cite{duvenaud2015convolutional,dai2016discriminative}, protein structure prediction~\cite{hamilton2017inductive}, and social network analysis~\cite{qiu2018deepinf}.
However, recent studies show that GNNs are vulnerable to adversarial manipulation, where carefully crafted instances are able to mislead machine learning models to make an arbitrarily incorrect prediction. Such vulnerabilities have raised great concerns when applying GNNs to security-critical applications. In particular,  different types of attacks targeting on GNNs by adding/deleting a small amount of edges within a target graph have been proposed to fool the node classification or subgraph classification tasks~\cite{dai2018adversarial,zugner2018adversarial,zugner2018meta}. These attacks are shown to be possible in real world scenarios. For example, as shown in Figure \ref{fig:eg}, a malicious user can bypass the malicious access detection system by linking himself with other legitimate users~\cite{manadhata2014detecting}. 

Detecting such adversarial attacks on GNNs involves several challenges.
First, such adversarial attacks focus on local graph properties and aim to create ``unnoticeable'' perturbations. Therefore, the manipulation of a small number of local edges is not obvious enough to be detected by traditional Sybil detection methods~\cite{bansal2016sybil}.
Second, existing defense/detection methods against adversarial behaviors on machine learning models are not easy to be applied to detect malicious attacks on GNNs for several reasons.
For instance, \emph{Robust generative models} are proposed to mitigate adversarial perturbation via denoising autoencoders and Generative Adversarial Networks (GAN) respectively~\cite{NIPS2014_5423,Meng2017MagNet,samangouei2018defensegan}; however, subtle adversarial perturbation in graph-structured data is hard to remove directly through generative models due to the discrete nature of graph data.
Third, the perturbations on a graph will show \textit{diverse} behaviors due to different factors, which makes it impossible to learn unified rules to identify them, and in-depth understanding of such adversarial behaviors is required. For instance, adding adversarial edges to a node of higher degree will induce different adversarial patterns (adversarial edges in this case are more likely to be outliers) compared with adding them to a node of low degree.

\begin{figure*}[ht]
    \centering
    \includegraphics[width=0.85\textwidth]{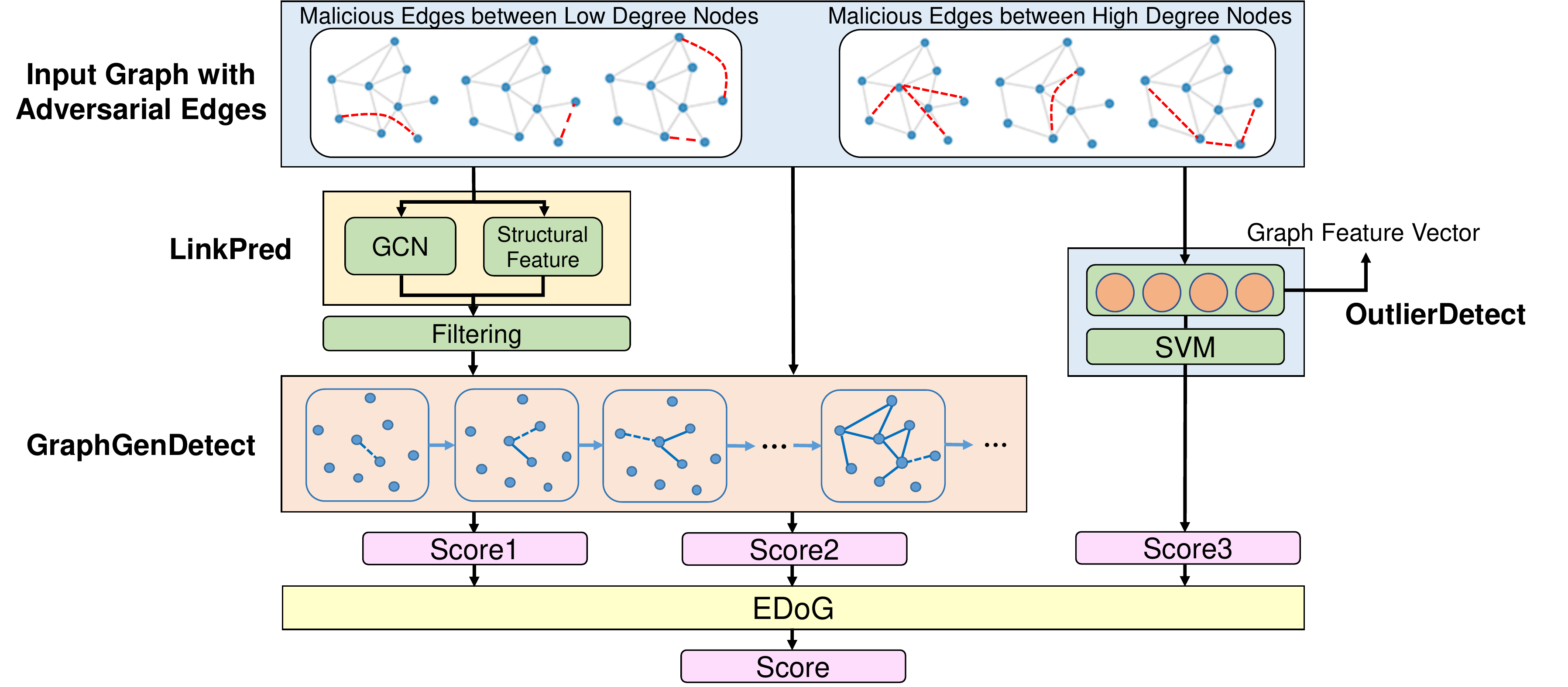}
    \caption{Illustration of the proposed general detection pipeline EDoG.}
    \label{fig:pipeline}
    \vspace{-3mm}
\end{figure*}

Given these challenges, in this paper 
we propose a general detection pipeline \ensembs as shown in Figure~\ref{fig:pipeline} to detect different unseen adversarial attacks in GNNs.
In particular, we propose several detection strategies as components of \ensembs, including Link Prediction based method (\linkps), Graph Generation based method (\graphgs), and Outlier Detection based method (\abnorms).
Here we mainly consider four types of attacks: (1) adding only one adversarial edge \cite{dai2018adversarial}; (2) the number of added adversarial edges is allowed to be up to the degree of a chosen target node, while these adversarial edges are all directly connected to the target node \cite{zugner2018adversarial}; (3) the number of added adversarial edges is allowed to be up to the degree of a chosen target node, while they are not directly connected to the target node \cite{zugner2018adversarial};(4) the number of added adversarial edges is allowed to be up to a certain percentage (e.g., 5\%) of the number of the edges in the original graph  \cite{zugner2018meta}.

We found that adversarial edges in the first type of attacks can be characterized by graph generative models which are trained with a large number of subsampled graphs. Therefore, we propose a novel graph generative model based detection method \graphgs to detect such adversarial edges. In particular, we first sample a number of subgraphs for training the generative model; and since the original graph contains a small number of adversarial edges, based on the \textit{union bound} these sub-graphs will not contain malicious edges with high probability. Such generative models can learn useful patterns from subgraphs and detect malicious edges.
In addition, we propose to use link prediction models to filter out suspicious edges coarsely to ensure that the generative models are trained on ``clean'' edges.

However, when a larger number of adversarial edges are allowed (i.e., the other three types of attacks), the sampled sub-graphs may contain more adversarial edges and therefore the trained graph generative model and link prediction model are not accurate enough to distinguish the added adversarial edges. In such scenarios, we found that the added adversarial edges are more likely to appear as ``outliers,'' while the single adversarial edge would not. As a result, we propose a list of intrinsic features that can be leveraged to perform outlier detection (\abnorms), together with \graphgs and \linkps.

By leveraging the intrinsic tradeoff between the stealthiness (hard to appear as ``outliers") and sparsity (hard to be sampled in subgraphs) of attacks, the proposed \ensembs is able to effectively identify adversarial edges effectively based on at least one of these properties.  
We conduct extensive experiments on three real-world graph datasets including one private transaction rule graph from a major company as well as two types of synthetic graphs (e.g., Erdos-Renyi and scale-free graphs) with controlled properties. We show that \abnorms can achieve detection AUC above 0.9 for the cases when the target nodes have high degree (larger than 10), which means that if we have knowledge about the attack type it is possible to achieve nearly perfect detection rate.
Even without the knowledge of attack types, our results show that the proposed general pipeline \ensembs outperforms other state-of-the-art adversarial edge detection methods~\cite{perozzi2016recommendation,rattigan2005case} significantly.
In addition, we also evaluate an adaptive attack in which the attacker has full knowledge of our detection pipeline and intentionally aims to bypass it during the attack. We show that the adaptive attack success rate remains very low given our detection pipeline. 
In summary, we make the following contributions: 
\begin{itemize}
    \item We propose a general light-weighted adversarial edge detection pipeline \ensembs on GNNs
    against the state-of-the-art GNNs based attacks. In particular, we evaluate \ensembs to detect against four adversarial attacks on GNNs without requiring knowledge of the attack strategies.
    \item We propose a novel graph generative model and a filter-and-sample framework to train the graph generative model for adversarial edge detection purpose.
    \item Based on several interesting observations, we provide a set of effective features on graphs that can be leveraged to perform outlier detection against unseen adversarial attacks as preprocessing.
    \item We conduct extensive experiments on Cora, Citeseer, transaction rule graph, and synthetic data with controlled properties to detect the state-of-the-art attacks, demonstrating the effectiveness of the proposed pipeline. The detection AUC of \ensembs can reach 0.8 without any knowledge of attack types and over 0.85 when we know the attack type. In both scenarios, the proposed \ensembs approach 
    outperforms other baseline methods.
    \item We evaluate \ensembs against a strong adaptive attack and show that it is difficult to bypass our detection pipeline.
\end{itemize}

\section{Background}

In this section, we will introduce background on neural networks, GNNs and the adversarial attacks on GNNs. 

A ``$c$-way classification task" in machine learning is a problem that given an input $x$ and $c$ classes, the model is required to predict which class the input belongs to. In order to deal with a $c$-way classification task, the output of a neural network is $p \in \mathbb{R}^c$ where the $i$-th value in $ p$ corresponds to the probability that the input belongs to class $i$. If we know the ground truth class $y$, we can evaluate the prediction by calculating the cross entropy loss between them:
\begin{equation*}
    L({p}, y) = -\log({ p}_y)
\end{equation*}
In order to generate a predicted class for the input, the model will take the class with the largest probability:
\begin{equation*}
    \hat{y} = \argmax_{y_0}~p_{y_0}
\end{equation*}

\subsection{Graph Neural Networks}
Graph Neural Networks (GNNs) are a class of nerual networks which processes graph data $G$. In general, $G=(V, E, X)$, where $V=\{v_1, v_2, \ldots\}$ denotes the set of nodes, $E=\{e_1, e_2, \ldots\},~e_i\in V \times V$ denotes the set of edges and $X=(\bm{x}_1, \bm{x}_2, \ldots, \bm{x}_{|V|})$ represents the feature vector of each node. 
Graph data is different from traditional machine learning data in that in addition to the node features, each node has relationship with its neighborhood indicated by the edges.
Therefore, a GNN model is usually designed such that each layer will consider the information in the neighbourhood of each node.
In particular, a GNN based model calculates an embedding vector $\bm\theta_u$ of each node $u\in V$ via iteratively aggregating information of itself and its neighbours:
\begin{equation*}
\bm{\theta}_u^{(k)} = f_k\big(\bm{x}_u, \bm{\theta}_u^{(k-1)}, \{\bm{x}_v, \bm{\theta}_v^{(k-1)}\}_{v \in \mathcal{N}(u)}\big),
\end{equation*}
where $\mathcal{N}(u)$ denotes the neighbours of $u$ in the graph. The node features $\bm{x}_u$ serves as the initial embedding $\bm{\theta}_u^{(0)}$. After we calculated the node embedding, we can use it in different tasks such as node classification and abnormal edge detection.

Many GNN models have been proposed such as Graph Convolutional Networks (GCN) \cite{kipf2016semi} and Structure2Vec \cite{dai2016discriminative} with impressive performance on various tasks. In this paper, we will focus on the GCN model. 
Let $\Theta^{(k)} = (\bm\theta_1^{(k)}, \bm\theta_2^{(k)}, \ldots, \bm\theta_{|V|}^{(k)})$ be the matrix of all node embedding vectors at step $k$. For GCN, the aggregation function is calculated as:
\begin{align*}
    \Theta^{(k)} &= \sigma\big(\hat{A} \Theta^{(k-1)} W^{(k)}\big)\\
    \hat{A} &= \Tilde{D}^{-\frac12} \Tilde{A} \Tilde{D}^{-\frac12}
\end{align*}
where $\Tilde{A} = A+I_N$, such that $A$ is the adjacency matrix and $I_N$ the identity matrix. $\tilde{D}$ is a diagonal matrix such that $\tilde{D}_{ii} = \sum_j\Tilde{A}_{ij}$. The function $\sigma$ is a non-linear activation function, and $W^{(k)}$ is the trainable parameters at the $k$-th layer.

\textbf{Node classification\quad}
In a node classification task over graph data, each node $v_i$ has a label $y_i \in \mathcal{Y}$, but we only have access to a small subset of the true labels, i.e., $L_\textit{train}: V_\textit{train} \rightarrow \mathcal{Y}$ where $V_\textit{train} = \{v_{i_0}, v_{i_1}, \ldots\} \subset V$ is the set of nodes for which we know the true labels. We also have a set of nodes $V_\textit{infer}$ with $V_\textit{train} \cap V_\textit{infer} = \emptyset$. Given $L_\textit{train}$, we would like to infer the labels of $V_\textit{infer}$. That is to say, we seek for a model to give a prediction $\hat{y}_i \in \mathcal{Y}$ to each of the node $v_i$, and we would like to maximize the classification accuracy, i.e., 
$\frac1{|V_\textit{infer}|} \sum_{v_i \in V_\textit{infer}} \mathbf{1}\{\hat{y}_i = y_i\}$.

When performing the node classification task, a GNN model $f$ first calculates the embedding $\bm\theta_u$ for each node in graph $G$, which will then be used to calculate the probability vector
 \begin{equation*}
\bm{p}_u = f(G, u) = \textrm{softmax}(W^\textit{out}\bm\theta_u)
 \end{equation*}
indicating the probability of each class that node $u$ belongs to. During the training process, the goal is to minimize the cross-entropy loss for the prediction of nodes in $V_\textit{train}$. During the evaluation process, the predicted class of each node $u$ is given by $\hat{y}_u = \argmax_y (\bm{p}_u)_y$.

\subsection{Adversarial Attacks on Graph-structured Data}
Recently, several studies \cite{dai2018adversarial,zugner2018adversarial,zugner2018meta} have shown that graph neural networks for node classification are vulnerable under adversarial attack. This means that a malicious attacker can modify the graph $G$ subtly into $G'$ before it is fed into the graph neural network such that the graph neural network will generate wrong classification results as desired by the attacker. \cite{dai2018adversarial,zugner2018adversarial} transform this setting into an optimization problem:
\begin{align*}
    \max_{G'} & \qquad L(f(G',v_t), y_t) \\
    \text{s.t.} & \qquad \mathcal{I}(G,G',v_t) = 1
\end{align*}
where $L(\cdot, \cdot)$ is the cross-entropy loss function between prediction vector of a node $v_t$ and its ground truth class $y_t$. $\mathcal{I}(G,G',v_t)$ is an equivalency indicator which judges whether the small modification between $G'$ and $G$ is reasonable. This indicator function may vary under different attack setting.

Note that this optimization goal cannot be solved by gradient-based approach because the constraint space is discrete - the value in the adjacency matrix of a graph can only be 0 or 1. In order to solve this problem, \cite{dai2018adversarial} defines the equivalency indicator such that the attacker is only allowed to add/delete one edge. The authors propose to parametrize the perturbation generator $G'=h(G, v_t)$ as a neural network and trains it with reinforcement learning \cite{sutton1998introduction}. In \cite{zugner2018adversarial} the equivalency indicator is defined such that the total amount of changed edges and node features is bounded. They propose a simple approximate model of GNNs on which they can analytically solve the optimization problem.

In \cite{zugner2018meta}, the optimization problem is defined differently so that it is not restricted to a single node, but over a set of nodes (usually the entire graph):
\begin{align*}
    \max_{G'} & \qquad \sum_{(v,y) \in V_{atk}} L(f(G',v), y) \\
    \text{s.t.} & \qquad \mathcal{I}(G,G',V_{atk}) = 1
\end{align*}
They solve the optimization problem by approximating the gradient of the adjacency matrix of $G'$ so that traditional gradient-based approach is employed.

\section{Threat Model and Detection Goal}

In this section, we will first introduce the threat model on graph neural networks. Then we will discuss our goal as detecting adversarial attacks against GNNs.

\subsection{Attack Model}
\label{sec:att-model}
Here we mainly focus on the node classification task, where GNNs are used to perform node classification on graph $G=(V,E,X)$. Based on the Kerckhoffs's theory~\cite{shannon1949communication}, we consider the strongest attacker who has whitebox access to the trained GNNs including the model architecture and parameters. 
The attacker aims to perform evasion attacks given a trained GNNs, and the strong attack assumption allows us to best evaluate the detection ability of EDoG.
{\bf {The attacker's goal}} is to change the predicted label of a \emph{target node} $v_{t}$ from the ground truth $y_{t}$ to the adversarial target $\hat{y'}_t $ ($\hat{y'}_t \neq y_t$) by manipulating the input test graph data in a subtle. 

There are mainly two categories of attacks on GNNs:
\begin{itemize}
    \item Feature attack: The attacker makes small modification to the feature vectors of nodes on the graph. In this attack the graph structure remains unchanged, i.e. $G'=(V,E,X')$.
    \item Structure attack: The attacker adds or delete a small number of edges in the graph. In this attack the node feature remains unchanged, i.e. $G'=(V,E',X)$.
\end{itemize}
An attacker may use either or both of the above approaches to attack GNNs. The feature attack is similar to the attack over other continuous data such as computer vision \cite{goodfellow2014explaining} where gradient-based optimization techniques can be used. Many works have been conducted against such kind of attacks. The structure attack, on the other hand, is a newly-proposed attack over discrete data structure. Hence, we mainly focus on the structure attack in this paper and assume that the node features are not changed.

Without loss of generality, here we mainly focus on attackers that add different numbers of malicious edges to the graph, which means $E \subset E'$. There are two reasons for such setting. First, adding edges is usually a cheaper  and more practical attack approach than deleting edges. For instance, in an undirected citation network an author can easily add edge by citing others in their own paper but cannot delete edges if their paper has been cited by others. Second, we empirically find that adding edges will yield higher attack success rate than deleting for attackers. As a result, adding edge attack is a more severe threat for learning tasks on graph structured data which we will mainly focus on. On the other hand, we can also leverage the inverse graph to analyze the attacks for deleting edges.

As for the structure attack, we consider the state-of-the-art attack strategies targeting on GNNs~\cite{dai2018adversarial,zugner2018adversarial}. 
Based on the number of allowed malicious edges and whether these malicious edges are directly connected to the target node of an attacker so as to make direct impact, we can categorize the attacks into three types as below. 

\begin{enumerate}
    \item \textbf{Single-edge attack}. This attack is proposed by \cite{dai2018adversarial} where an attacker is allowed to add only one malicious edge to the graph to perform stealthy attack, i.e. $|E' \backslash E| = 1$.
    \item \textbf{Multi-edge direct attack}. This attack is proposed by \cite{zugner2018adversarial} where an attacker is allowed to add several edges to the graph. The maliciously added edges would be connected to the target node and the number of malicious edges should not exceed the degree of the target node. That is, $|E'\backslash E| \leq \text{deg}(v_t)$ and for any edge $\forall e(i,j) \in E' \backslash E$, $v_t \in e(i,j)$.
    \item \textbf{Multi-edge indirect attack}. This attack is similar to the Multi-edge direct attack except that the added malicious edges are not directly connected to the target node. That is, $|E'\backslash E| \leq \text{deg}(v_t)$ and for edge $\forall e(i,j) \in E' \backslash E$, $v_t \notin e(i,j)$.
\end{enumerate}

\begin{figure}[t]
    \centering
    \includegraphics[width=0.7\linewidth]{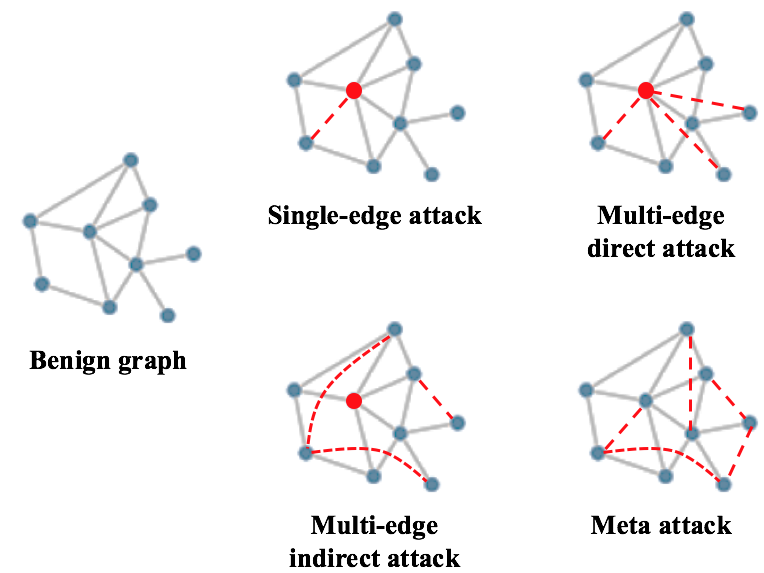}
    \caption{Examples of four state-of-the-art of attacks we considered in this paper. The red node refers to the target node in the attack and the red dashed lines are the maliciously added edges which we aim to detect.}
    \label{fig:attack}
\end{figure}

In addition, \cite{zugner2018meta} proposes another structure attack which is known as meta attack. In meta attack, the attacker does not have a target node but focuses on the entire graph. The attacker's goal is to make the GNN model to give wrong classification result on as much nodes as possible. They allow the number of added edges to be up to 5\% of the total number of existing edges in the graph. Hence, we have the fourth type of attack considered in the paper:
\begin{enumerate}
    \item[4)] \textbf{Meta attack}. Here the attacker can add at most 5\% of malicious edges, i.e. $|E'\backslash E| \leq 5\% \times |E|$.
\end{enumerate}
Examples of the four types of attacks are shown in Figure \ref{fig:attack}.

\subsection{Goals of Adversarial Edge Detection}
In order to deal with the aforementioned threat, we aim to propose a general pipeline to detect the maliciously added edges $|E' \backslash E|$. 
Our \textbf{detection goal} is as follows. Suppose the defender is provided with a graph $G' = (V, E', X)$. The defender knows that some of the edges may be added maliciously by the attacker as above, and he/she does not have any other information about the attack (e.g. the degree of target nodes or the attack strategies). 
The goal is to determine which edges in $E'$ are likely to be malicious. On the high level, the defender will calculate a score $s_j$ for each edge $e_j \in E'$ indicating how likely the edge may be a malicious one.
After calculating this score for chosen edges, he/she can either identify adversarial edges directly (by setting a proper threshold) or set priorities for further inspection of ``suspicious" edges.

In particular, we would like the detection pipeline to satisfy the following properties:
\begin{itemize}
    \item The defender only sees the modified graph $G'$ and does not have information about the original graph $G$. Otherwise the task would become trivial.
    \item The pipeline should work without any information about the attack, such as the attack strategies or the target node. We will also show later that with such auxiliary information the pipeline may achieve a better performance as ablation studies for understanding purpose.
    \item The detection pipeline should be general against different attacks. EDoG should be able to generalize and detect malicious edges for 1) different types of graphs, 2) various of attack algorithms and 3) different target nodes with different degrees.
\end{itemize}

\begin{figure*}
    \centering
    \includegraphics[width=0.8\linewidth]{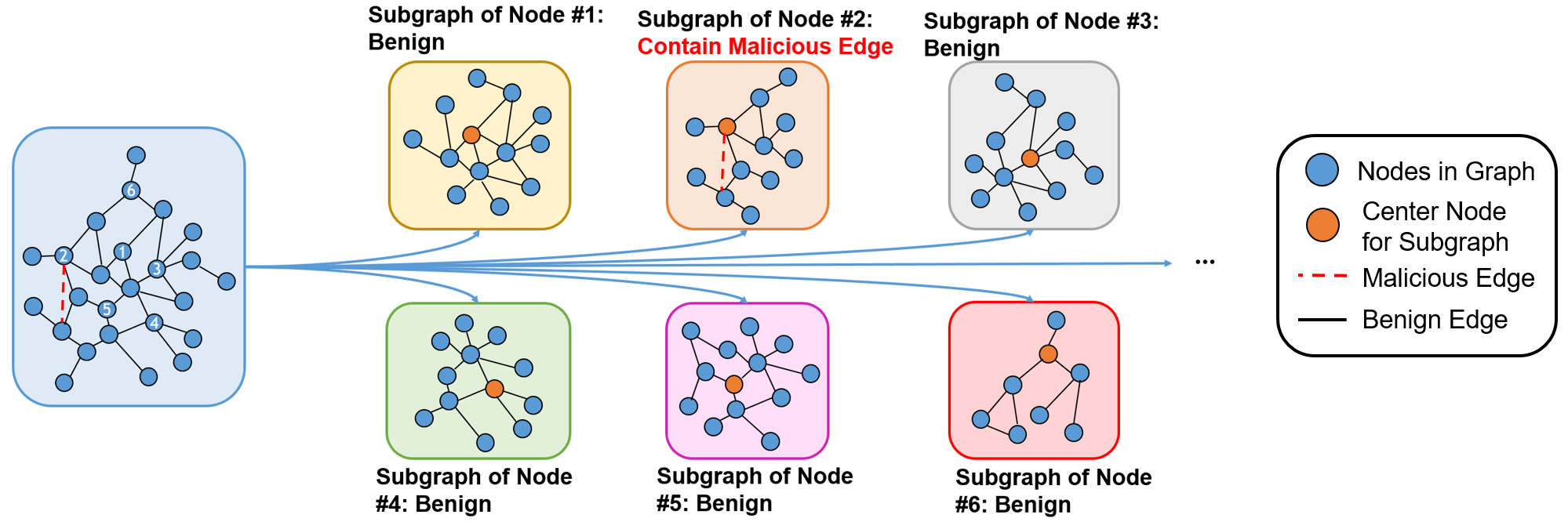}
    \caption{An illustration of sampling subgraphs to train the graph generative model to capture benign graph structure properties. It is shown that most sampled graphs will not contain malicious edges when only one adversarial edge is added.}
    \label{fig:sampling}
\end{figure*}

\section{Analysis of Adversarial Attacks on Graphs }
In this section, we will first summarize the findings and insights about adversarial edge properties of different types of attacks. We then provide the high level overview and intuitions of our proposed detection approaches, as well as the final detection pipeline EDoG.

\subsection{Malicious Links between Low-degree Nodes: Link Prediction and Generative Model}
\label{sec:intuition-lp}
Here we consider the attack model where the attacker only adds a small number of adversarial edges, which is a sufficient condition for target nodes with low degree. This is because for all the attacks we considered, the maximum number of allowed adversarial edges is up to the degree of target node.
When an attacker only adds a very small number of edges to ensure the perturbation is ``unnoticeable", we will first assume that the perturbation of malicious edges is small enough to be neglected---any algorithm, except for the target GNN based models, applied over the malicious graph $G'$ will return approximately the same result as if it is applied over $G$. Thus, an intuitive approach is that we can train a link prediction model using $G'$. Ideally, the link prediction model would behave as if it were trained using $G$, so it will predict the node pairs with edges as high scores and the pairs without edges as low scores. The malicious edges, however, do not exist in the benign graph. So the scores of them will be low. Thus, after applying the link prediction algorithm we can check the scores for the edges in $G'$. Those with low scores should be considered likely to be malicious.
In this link prediction based approach, selecting features that would focus more on global structures are more helpful, and therefore we propose to use the GCN based structural features for prediction. We will discuss the used features in detail in the next section.

In addition, in order to better capture the global structure information of the original graph data, we also propose generative models to better approximate the original link distribution. A key step for training generative models is to ensure that the data are strictly \emph{clean} (not malicious) to avoid being ``poisoned".
One approach for reducing the effect of malicious edges would be to sample subgraphs from the large graph, as illustrated in Figure \ref{fig:sampling}. The intuition is that if we randomly sample many small sub-graphs from a large graph, each edge will only appear in a small proportion of the subgraphs with high probability by union bound. Thus, most subgraphs will contain no malicious edges while preserving the information of the original graph. For example, in our experiment we find that if we sample the graphs by extracting the two-hop neighbour for each node, it turns out that the more than 99\% of subgraphs do not contain malicious edges in single-edge attack on average on the dataset we use, and more than 90\% for multi-edge attack and meta-attack. Therefore, we can train generative models over the subgraphs to learn a good distribution approximation of the original graph, and then leverage this to detect abnormal edges. Note that the naive link prediction algorithm is not suitable for training over subgraphs in two aspects: first, many link prediction algorithms are based on feature extraction over the entire graph; second, link prediction algorithms tend to have relatively small model capacity to capture the pattern of entire sub-graphs. Therefore, we can only train the proposed graph generative models based on deep neural networks on subgraphs. Based on the generative models, we will be able to identify which edges are the least likely to be generated and these edges are highly likely to be malicious. We find that such generative models are good at discovering patterns from subgraphs and thus detect malicious edges.


\subsection{Malicious Links between High-degree Nodes: Outliers}
\label{sec:intuition-od}

\begin{figure}[t]
    \centering
    \begin{subfigure}[t]{0.45\linewidth}
        \centering
        \includegraphics[width=0.9\linewidth]{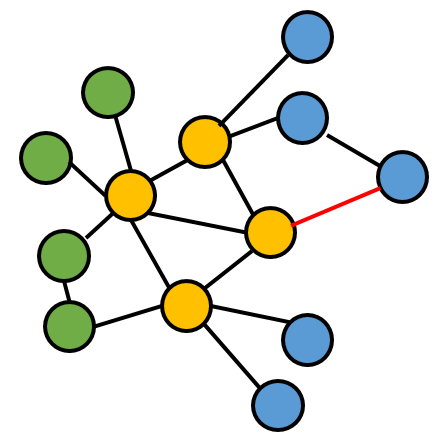}
        \caption{Add one edge.}
    \end{subfigure}
    \begin{subfigure}[t]{0.45\linewidth}
        \centering
        \includegraphics[width=0.9\linewidth]{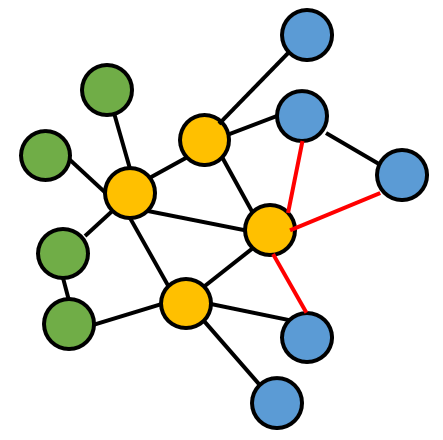}
        \caption{Add three edges.}
    \end{subfigure}
    \caption{An example of of ``collective power of malicious edges". When only one malicious edge is added (left), it should be easily detected be the graph generative model; but when more malicious edges are added (right), all the malicious edges will appear to be benign.}
    \label{fig:collect}
    \vspace{-3mm}
\end{figure}

It turns out that both link prediction and graph generation approach may sometimes fail when applied to multi-edge direct attack, especially when the node degree is large. We attribute this to a principle of the \emph{collective power of malicious edges} which can be understood as: when there are many malicious edges connecting to one node, they confirm the legitimacy of each other mutually. We show an example as in Figure \ref{fig:collect}. Suppose one node is originally connected to three nodes in class 1. If an attacker adds just one malicious edge that connects it to a node in class 2, this edge will seem abnormal and is easily detected. However, if the attacker instead adds three malicious edges in class 2, the legitimacy of each malicious edge will be supported by the rest, and they will all be judged to be benign. 

Under this circumstance, the neighbours of the target node in $G'$ should contain a number of different classes (e.g. 50\% in class 1 and 50\% in class 2), while the classes of other nodes' neighbors are usually quite uniform. As a result, we may calculate several edge features indicating the information of the neighbourhood of edges, e.g. number of different classes appearing in the neighbourhood of the edge. We could expect that the feature vector of a malicious edge would be different from those of benign edges. Therefore, we could build an outlier detection model over all the edges and consider the outliers as malicious.

\section{\ensembs: Adversarial Edge Detection}
\label{sec:method}
In this section, we will introduce the proposed primitives for adversarial detection considering different scenarios, followed by the proposed general detection pipeline EDoG.

\subsection{\linkp(\linkps)}
A direct link prediction algorithm is first considered to integrate given its simplicity and effectiveness. For example, in \cite{perozzi2016recommendation} the authors calculate five features (as we will discuss in Section \ref{sec:setup}) for each node pair $(u,v)$ and use the feature vector to judge whether the node pair should be connected. However, we can do more than that---using GNN, we may discover some latent feature space in addition to the features used in link prediction.
In practice, we first adopt a two-layer graph convolutional network to calculate the node embedding vector $\bm\theta_u$ for each node. Then for each pair node $(u, v)$ we use a bilinear hidden layer to calculate a hidden feature 
\begin{equation*}
f_{u,v}^\textit{GNN} = \sigmoid(\bm\theta_u^\intercal W_\textit{edge} \bm\theta_v)
\end{equation*}
where $\sigmoid(\cdot)$ is the sigmoid function. This feature is appended to the feature vector of each node pair to form a 6-dimensional vector and then passed through a logistic regression layer. The entire model is trained end-to-end. The inference process is similar to \cite{perozzi2016recommendation}---we calculate the score for all existing edges, and the edges with low scores are likely to be malicious.

\subsection{\graphg(\graphgs)}

\begin{algorithm}[bt]
\caption{Algorithm for predicting how likely each node pair in $E_{target}$ should be linked in the graph $G=(V,E,X)$. $f_{gen}$ in the algorithm refers to the Graph Generation Model which takes a graph as input and output the the score indicating the probability of link for each unlinked node pair in the graph.}
\label{alg:ggd}
\begin{algorithmic}[1]
\Procedure{EdgeLinkProbs}{$f_\textit{gen}$, $G$, $E_\textit{target}$}
\State $\textit{LinkProbs} \gets [~]$
\For {$e$ \textbf{in} $E_\textit{target}$}
\State $\textit{LinkProbs}[e] \gets [~]$
\EndFor
\State $\pi = \textit{permutation}(|E|)$
\State $E_0 = \emptyset$
\For {$t$ \textbf{in} $0,1,\ldots,|E|-1$}
\For {$e$ \textbf{in} $E_\textit{target} \backslash E_t$}
\State $\textit{LinkProb}$ = $f_\textit{gen}(V,E_t,X)[e]$ 
\State $\textit{LinkProbs}[e] \gets \textit{LinkProbs}[e] + [\textit{LinkProb}]$
\EndFor
\State \textit{/* Add one edge in each iteration. */}
\State $E_{t+1} \gets E_t \cup {E[\pi_t]}$ 
\EndFor
\For {$e$ \textbf{in} $E_\textit{target}$}
\State $\textit{LinkProbs}[e] \gets \textit{average}(\textit{LinkProbs}[e])$
\EndFor
\State \Return $\textit{LinkProbs}$
\EndProcedure
\end{algorithmic}
\end{algorithm}

\begin{algorithm}[bt]
\caption{Algorithm for using Graph Generation model $f_{gen}$ to detect malicious edges on graph $G'=(V,E',X)$.}
\label{alg:infer}
\begin{algorithmic}[1]
\Procedure{MaliciousEdgeDetect}{$f_{gen}, G'$}
\State \textit{/* Sample sub-graphs. */}
\State $G'_0,G'_1,\ldots,G'_n = \textit{Sample}(G')$
\State $LinkProbs \gets []$
\For {$e$ \textbf{in} $E$}
\State $LinkProbs[e] \gets []$
\EndFor
\State \textit{/* Enumerate through sub-graphs. */}
\For {$G'_i$ \textbf{in} $G'_0, G'_1,\ldots,G'_{n-1}$}
\State $SubLinkP \gets$ EdgeLinkProbs($f_{gen}, G'_i, E$) 
\For {$e$ \textbf{in} $E'_i$}
\State $LinkProbs \gets LinkProbs + SubLinkP[e]$
\EndFor
\EndFor
\State \textit{/* Probability of being malicious. */}
\State $MalProbs \gets []$
\For{$e$ \textbf{in} $E'$}
\State $MalProbs[e] \gets 1 - \textit{Average}(LinkProbs[e])$
\EndFor
\State \Return $MalProbs$
\EndProcedure
\end{algorithmic}
\end{algorithm}


Following the intuition in \ref{sec:intuition-lp}, we propose to train a generative models to capture the complex structure of different subgraphs.
There have been several works on graph generative models~\cite{simonovsky2018graphvae,You2018Graph,you2018graphrnn,Bojchevski2018NetGAN,ding2018semi}. However, most of the existing generative models aim to generate as diverse as possible graphs from the training set. On the other hand, our goal is to leverage the generative model to predict which edges are more likely to be generated, which means we hope to preserve the properties of original graph. Hence, we aim to design a generative model that could discover the graph structural pattern and does not need to include much diversity in the output.
As a result, we propose a deep graph generation model inspired by \textbf{sequence generation} approaches~\cite{sutskever2014sequence} -- the model will generate the edges one-by-one to construct the entire graph. We train the generative model based on randomly sampled subgraphs, and apply the trained model to predict which edges in $E'$ are the least likely to be generated. These edges with low generative likelihood would be considered to be malicious.

Our model, denoted as $f_{gen}$, will take as input a graph $G=(V,E,X)$ and output a probability distribution indicating that if we are going to add one edge into the graph, which node pair is likely to be added. That is to say, the output of $f_{gen}$ is a probability distribution over all the node pairs that have not been connected yet:
\begin{equation*}
    e^{(t)} \sim P\big[ (u,v) \big| (u,v) \notin E\big]
\end{equation*}
Based on this model, we can generate a graph given only node information $(V,X)$ in a step-by-step procedure. In particular, we start from an empty edge set $E^{(0)}$ and gradually add edges to the set. At each time step $t$, we input $(V,E^{(t-1)},X)$ into the $f_{gen}$, sample a new edge using the output probability distribution and add the new edge into the edge set to get $E{(t)}$.

In practice, we use a GCN with bilinear output layer to calculate the probability. We first apply a two-layer GCN to calculate the embedding vector $\bm\theta_u^{(t-1)}$ for each node $u$ at time $t-1$. Then, for each node pair $(u,v)$, we apply a bilinear function $s_{uv}^{(t-1)} = (\bm\theta_u^{(t-1)})^\intercal W \bm\theta_v^{(t-1)}$ to calculate the score. In order to determine which edge will be generated at the next time step, we take softmax of the scores of all the node pairs which have not been connected to calculate the probability as:
\begin{equation*}
    P\big[ (u,v) \big] = \text{softmax}\big(\big\{s_{uv}\big|(u,v) \notin E^{(t-1)}\big\}\big)
\end{equation*}

After training the generative model $f_{gen}$, we aim to calculate:\textit{ given a graph $G$ and a set of node pairs $E_{target}$, what is the likelihood ratio that there exists an edge between each node pair in $E_{target}$?}

The detailed algorithm is shown in Algorithm~\ref{alg:ggd}. In particular, given a subgraph $G_i$ we will 1)
randomly generate a permutation to get an order for edges. Following this order, we are going to start with an empty edge set and add one edge to the graph at each time step;
2) at each time step $t$, feed in the current adjacency matrix $A^{(t)}$ and node feature $X$ to model $f_{gen}$ and calculate the scores $s_{uv}$ for the desired node pairs in $E_{target}$; 3) calculate the final score of an edge as the average of the scores which we calculate in all the time steps. 

Using Alg.~\ref{alg:ggd}, we can infer which edges are likely to be malicious in graph $G'$ as in Alg.~\ref{alg:infer}. During this inference stage, we will first sample sub-graphs from $G'$ and calculate the probability of links for edges in each sub-graph. Then the score of edges in the original graph can be calculated by taking the average accordingly. A small score means the edge is unlikely to be generated, and therefore it is likely to be malicious.

We will use gradient-based method to minimize a training loss iteratively in order to train the model $f_{gen}$.
At each training step, the process for calculating the training loss is shown in Alg~\ref{alg:train}. Similar to the inference stage, we first sample sub-graphs from $G'=(V,E',X)$. However, in order to train the model, we not only calculate the probability of the linked node pairs (i.e. edges in sub-graphs) but also calculate the probability for a set of unlinked node pairs $E_{nonexist}$. This set is uniformly sampled from the unlinked node pairs in the graph and we sample a different set in different training step. We let the set size be $|E_{nonexist}| = |E'|$. Hence, the training loss at each time step is a binary cross entropy loss over all the node pairs. The ground truth label is 1 if the node pairs is linked in the original graph and 0 otherwise. After calculating the loss, we can apply the gradient-based optimizer to update the model parameters in $f_{gen}$.

\begin{algorithm}[bt]
\caption{Algorithm for training the Graph Generation model $f_{gen}$ given a graph $G'=(V,E',X)$. $E_{nonexist}$ is a randomly sampled set of node pairs such that no edge exists among each node pair. }
\label{alg:train}
\begin{algorithmic}[1]
\Procedure{ModelTrainingLoss}{$f_{gen}, G', E_{nonexist}$}
\State \textit{/* Sample sub-graphs. */}
\State $G'_0,G'_1,\ldots,G'_n = \textit{Sample}(G')$
\State $loss \gets 0$
\State $E_{all} = E' \cup E_{nonexist}$
\For {$G'_i$ \textbf{in} $G'_0, G'_1,\ldots,G'_{n-1}$}
\State $SubLinkP \gets$ EdgeLinkProbs($f_{gen}, G'_i$, $E_{all}$)
\For {$e$ \textbf{in} $E_{all}$}
\State $label = e\textit{ is an existing edge ? }1:0$
\State $loss \gets loss + \textit{cross\_ent(label,SubLinkP[e])}$
\EndFor
\EndFor
\State \Return loss
\EndProcedure
\end{algorithmic}
\end{algorithm}


\subsection{Filtering for \graphg}
The graph generation model indeed has a strong capacity in learning patterns from sub-graphs. However, the existence of malicious graphs may still harm the performance of \graphg because our graph generation model would treat the pattern of malicious edges as benign and learn it well, leading the algorithm to be unstable. Therefore, we propose an effective \emph{filtering} process, i.e. we can first filter away a proportion of edges in the original graph which seems to be suspicious and train our generation model on the filtered graph. If the malicious edges are indeed filtered away, most of the sampled graphs should be benign during the training process and we can expect that our trained generative model will capture the structure properties of benign subgraphs.

In order to decide which edges are likely to be malicious and should be filtered, we can use algorithms introduced before to evaluate the maliciousness score for each edge. In practice, we find our \linkp approach a good way to filter away edges, as it is a stable algorithm and could reach acceptable performance to filter away malicious edges in most cases. We denote this approach as \linkp+\graphg. That is to say, we first apply \linkp over $G'$ and to get a score for each edge indicating how likely it is malicious. Then we will remove the top $k$ edges with highest malicious scores, getting $G'_\textit{filter}=(V,E'_\textit{filter},X)$. Thus, we can train our \graphg model on $G'_{filter}$ and finally use the trained model to detect the malicious edges over $G'$.

\subsection{\abnorm(\abnorms)}
Based on the intuition in Section~\ref{sec:intuition-od}, we propose the \abnorm to further identify adversarial edges mainly for high-degree victim nodes.
The intuition of this approach comes from an observation on the attacker behaviour during direct structure attack: in order to make the target victim node be misclassified into another class (say class 1), the attacker tends to add many edges between the target nodes with  nodes that belong to class 1. Hence, the class distribution of the target node's neighbours can be quite diverse considering its previous connections. In contrast, the classes in a benign node's neighbourhood should be quite uniform.
Inspired by this phenomenon, we propose a novel outlier detection model for the edges based on the class distribution of the neighbourhood nodes of an edge. 
In particular, we calculate the following features for each node:
\begin{itemize}
    \item The number of different classes of the neighbourhood nodes.
    \item Average appearance time of each class in the neighbourhood nodes.
    \item Appearance time of the most frequently appeared class in the neighbourhood nodes.
    \item Appearance time of the second most frequent class in the neighbour (0 if only one class is in the neighbourhood).
    \item Standard deviation of the appearance time of each class in the neighbourhood.
    \item Logarithm of the betweenness centrality \cite{freeman1977set} of each node in the graph.
\end{itemize}
Note that we do not have ground truth classes information for most nodes, so we would first fit a GNN over the graph $G'$ and use the prediction as the label.
For each edge, we calculate the above features for both nodes and concatenate them together, constructing a 10-dimensional feature vector. We then train a one-class SVM with the RBF kernel over these edge feature vectors to detect the outliers. The trained model will calculate a score for each edge indicating its abnormality. The larger the value is, the more likely that the edge is a malicious one.

\subsection{General Pipeline for Detection - \ensemb} 
The primitive approaches we propose above focus on different attacking scenarios.
In particular, \linkp+ \graphg and \graphg approach work well in most cases, except for the case when many malicious edges are connected to a single node (target node with high degree). On the other hand, \abnorm is very good at detecting such kinds of attacks with target nodes of high degree.
Therefore, similar to \cite{chen2018bootstrap}, in which a strong model is fit over graph data, we need to design a unified framework for the overall detection on adversarial edges.

Based on the node degree information, our final pipeline, namely Edge Detection of Graph (\ensemb), is shown in Figure~\ref{fig:pipeline}. It is an aggregated model which averages the output of \linkp + \graphg, \abnorm and \graphg. In particular, we apply the three approaches over the graph and get their prediction score for each edge. Then for edges between high-degree nodes (in practice we choose the criteria that the sum of degree of the two nodes is larger than 6), we use the average of scores from the three approaches; otherwise, for low degree nodes we only use the average score of \linkp + \graphg and \graphg. The advantage in doing this is: first, \graphg and \linkp + \graphg perform not very well at attacks with high node degree by Multi-edge direct attack, and incorporating it with \abnorm significantly improve the detection performance; second, the \graphg-based approach performs well but sometimes not very stable, so an aggregated model could help improve its robustness significantly.


\section{Experimental results}
\label{sec:exp}



In this section, we will first introduce the dataset and evaluation metrics we use, followed by the attack models, and performance analysis of the proposed detection methods.

\subsection{Experimental Setup}
\label{sec:setup}

\textbf{Datasets and evaluation metrics.}
We evaluate our detection model on two public citation networks, one private transaction rule graph from a major company, and two types of synthetic dataset with controlled properties. The benign accuracy of our model on the datasets are shown in Appendix~
\ref{sec:app-benign-acc}. We hope that such variety of  datasets can demonstrate the flexibility of our approach.

The real-world network datasets we use are Cora~\cite{mccallum2000automating} and Citeseer~\cite{giles1998citeseer}. These two datasets are citation networks where a node represents a research paper and an edge represents a citation between two papers. Cora has 2,708 nodes and 5,429 edges; Citeseer has 3,327 nodes and 4,732 edges. Each node contains a bag-of-word feature vector and has a ground truth label indicating which type of paper it is. During the training process, only a small part of node labels are available (140 for Cora and 120 for Citeseer). The model is trained to determine what are the labels for the other nodes. Cora is a 7-way classification task and Citeseer is 6-way. This setting is commonly used in the task of node classification on graph data \cite{kipf2016semi,dai2018adversarial,zugner2018adversarial}.

We also evaluate \ensembs on a private transaction rule graph from a major company. In this rule graph (denoted as Rule), each node represents a rule in the system and each edge represents a relationship that two rules are frequently co-triggered. We extract a subgraph with 719 nodes and 3,375 edges. Each node has a related feature vector, including information such as  author ID and its application in the system. The node classification task is to determine whether a rule is a temporary test rule or not.

For experiments on synthetic graphs with controlled properties, we generate two Erdos-Renyi graphs~\cite{erdHos1959random} and one scale-free graph~\cite{albert2002statistical}. The two Erdos-Renyi graphs are $G_{n,p_1}$ and $G_{n,p_2}$ with $n=1000$ and $p_1=\frac{\ln n}{n}$, $p_2=\frac{2\ln n}{n}$. The scale-free graph is generated using Barabasi-Albert algorithms, with 1,000 nodes and parameter $m=1$.
We would like to assign node features and node labels that are related with the graph structures. Therefore, given a synthetic graph we first assign a 20-dimensional random feature $e_u$ to each node $u$. Then we let the node features to correlate with its neighbours by repeat:
\begin{equation*}
    e_u = \sum_{v\in \mathcal{N}(u)} e_v, \qquad e_u = e_u / ||e_u||_2
\end{equation*}
where $\mathcal{N}(u)$ is the neighboring nodes of $u$. After repeat the process several times (in practice we repeat 3 times), we can get a hidden feature vector $e_u$ which is related with graph structures. The final node feature vector $x_u$ is a discrete random binary vector, the probability that the $i$-th bit of $x_u$ equals 1 is:
\begin{equation*}
    Pr[x_u^{(i)} = 1] = sigmoid(e_u^{(i)})
\end{equation*}
And the label of $u$ is $y_u = 1$ if $\sum_i x_u^{(i)} > 0$ and otherwise $y_u = 0$. The node classification accuracy can reach around 80\% for ER graphs and around 75\% for scale-free graphs. 

For each dataset and attack approach, we randomly pick several target victim nodes and perform the state-of-the-art attacks to generate malicious edges. Then we perform the detection method on the new graphs and check whether the malicious edges can be detected without attack information. The evaluation metric is the Area Under ROC Curve (AUC), which is a commonly used metric to verify the performance of a detection method. 

\textbf{Attack strategies.}
\begin{table}[t]
    \centering
     \caption{Degree of the selected target victim nodes. We aim to cover diverse target nodes with a large range of degrees.}
     \label{tab:deg}
    \begin{tabular}{c|c|c|c}
        \toprule
         & Cora & Citeseer & Rule \\
         \hline
         Single-edge & \makecell{1,2,3,4,5,\\6,6,7,8,10} & \makecell{1,2,3,3,4,\\4,5,6,7,9} & \makecell{1,2,3,4,4,\\5,5,6,7,10} \\
         \hline
         \makecell{Multi-edges\\direct} & \makecell{1,2,3,4,4,6,\\7,8,8,10,12,\\14,12,13,15,31,\\19,32,16,17} & \makecell{1,2,2,3,4,5,\\6,6,7,8,9,10,\\11,12,13,15,\\16,17,18,20} & \makecell{1,2,3,4,5,6,7,\\8,9,10,11,12,\\14,14,15,16,\\17,18,19,20} \\
         \hline
         \makecell{Multi-edges\\indirect} & 3,4,14,12,13,17 & 1,4,6,8,13,17 & 3,6,8,11,11,15\\
         \hline
     \end{tabular}
\end{table}
We evaluate the attack strategies as introduced in Section \ref{sec:att-model}.
We only choose the target nodes that are successfully attacked and randomly sample the target nodes with different degrees.
We observe in practice that Multi-edges direct attack is the most successful attacking model, followed by Single-edge attack and finally Multi-edges indirect attack. Therefore, we selected 20, 10 and 6 target nodes respectively for these three attack methods on real-world data.
For synthetic data, we simply pick two target nodes, one with the smallest degree and the other with the largest degree. The selected target node degrees are shown in the Table \ref{tab:deg}. For the meta-attack, we follow the same setting as in the paper\footnote{\url{https://github.com/danielzuegner/gnn-meta-attack}} and add 5\% malicious edges to the graph. We use a standard 2-layer GCN for the classification tasks and the training setting is the same as in \cite{dai2018adversarial,zugner2018adversarial,zugner2018meta}.

\begin{table*}[th]
    \centering
    \caption{The average AUC of EDoG and baselines against different attacks without knowledge of the attack type.}
    \label{tab:overall}
    \begin{tabular}{c|c|c|c|c}
        \toprule
        & Single-edge attack & Multi-edges direct attack & Multi-edges indirect attack & Meta attack \\
        \hline
        \bls & 0.5989 & 0.4293 & 0.3499 & 0.4759 \\
        \hline
        \blbs & 0.8502 & 0.6214 & 0.7985 & 0.6964 \\
        \hline
        \ensembs & \bf 0.8610 & \bf 0.7551 & \bf 0.8288 & \bf 0.7277 \\
        \hline
    \end{tabular}
    \vspace{-3mm}
\end{table*}

\textbf{Baseline Detection Approaches.} We compare the proposed EDoG with two state-of-the-art detection approaches as below. We will show the performance of other traditional metrics in 
Appendix~\ref{sec:app-other-metric}.
\subsubsection{\bl Approach (\bls)}
Our first baseline is the anomaly link detection approach as proposed in \cite{perozzi2016recommendation}. 
This approach trains a link prediction-based algorithm for detection. For each node pair $(u,v)$ it calculates five features, including the similarity of neighbours, the number of common neighbours, the distance between two nodes, the preferential attachment of two nodes and the similarity of the node features of $u$ and $v$. Then it trains a logistic regression classifier over the feature vectors of the node pairs, where the ground truth label of node pairs with edge is 1, otherwise 0. After the model is trained, we calculate the probability of link for each existing edge. The smaller the probability is, the more likely that edge is a malicious one. 

\subsubsection{\blb Approach (\blbs)}
Another baseline is an anomaly link discovery method described in \cite{rattigan2005case}, in which they used a high-order heuristic named Katz index \cite{liben2003link} to measure the connectivity of each node pair $(u,v)$ as
\begin{equation*}
Katz{(u,v)} = \sum_{l=1}^{\infty}{\beta^l|walks^{l}(u,v)|}
\label{eqn:katz}
\end{equation*}
where $\beta$ is damping factor to assign more weight to shorter walks and $walks^{l}(u,v)$ is the set of random walks with the length of $l$ between $u$ and $v$.  The intuition behind this measure is that compared with the normal links, malicious links often have small Katz index values. Then we can calculate the Katz index for each node pair and used the softmax function for normalization. After that, the value assigned for each node pair is bounded between 0 and 1, and we use this value as the probability for the existence of the link between the nodes. The smaller the probability is, the more likely that edge is a malicious one.

\textbf{Implementation Details.}
We implement all the detection models in Pytorch~\cite{paszke2017automatic} except for the \bls and \blbs and \abnorm where we use the scikit-learn toolkit~\cite{scikit-learn} for logistic regression and one-class SVM. In order to sample subgraphs from the original graph, we iterate through each node and extract its two-hop neighbourhood as a subgraph. Thus, we can get a set of subgraphs whose cardinality equals the number of nodes in the original graph. For \linkp, we train it for 500 epochs using SGD optimizer with learning rate of 0.01 and we sample among node pairs which are not connected so that the number of positive and negative labels is the same.
For \abnorm we fit a one-class svm with radial basis function kernel. For \graphg we train it using Adam optimizer~\cite{kingma2014adam} for 15 epochs with learning rate 0.001. We observe that the detection result of the \graphg model is reasonable fast (e.g. 5 epochs can be enough) while the result can be non-stable. Therefore, during test time we evaluate the trained model from the 6th to 15th epoch and take the average scores as the final prediction. For filtering model \linkp + \graphg, we filter away 50\% of the edges using the result of \linkp.

\begin{table*}[htbp]
    \centering
    \caption{AUC of the our detection approaches compared with baselines when we have knowledge of the attack type.
    }
    \vspace{-2mm}
    \label{tab:result-known}
    \begin{tabular}{c|c|c|c|c|c|c|c|c|c|c}
        \toprule
        & & \multicolumn{2}{c|}{Single-edge attack} & \multicolumn{4}{c|}{Multi-edges direct attack} & \multicolumn{2}{c|}{Multi-edges indirect attack} & Meta attack \\
        \cline{2-11}
        & Node degree & $[1,5]$ & $(5,+\inf)$ & $[1,5]$ & $(5,10]$ & $(10,15]$ & $(15,+\infty)$ & $[1,10]$ & $(10,+\inf)$ & - \\
        \hline
        \multirow{3}{*}{Cora} & \bls & 0.6745 & 0.4331 & 0.5363 & 0.4312 & 0.4372 & 0.4302 & 0.5384 & 0.4350 & 0.4820 \\
        \cline{2-11}
        & \blbs & \bf 0.9213 & 0.8282 & 0.7965 & 0.5727 & 0.3984 & 0.3946  & 0.8262 & 0.7562 & 0.6189\\
        \cline{2-11}
        & (\linkps + \graphgs) $\otimes$ \abnorms $\otimes$ \ensembs & 0.9088 & \bf 0.9110 & \bf 0.8319 & \bf 0.7378 & \bf 0.9144 & \bf 0.9123 & \bf 0.8368 & \bf 0.8414 & \bf 0.6884 \\
        \hline
        \hline
        \multirow{3}{*}{Citeseer} & \bls & 0.7942 & 0.5806 & 0.4853 & 0.2696 & 0.2726 & 0.2744 & 0.4229 & 0.3715 & 0.3956 \\ 
        \cline{2-11}
        & \blbs & 0.7677 & 0.6482 & \bf0.6049 & 0.4004 & 0.2851 & 0.2490 & 0.6401 & 0.6124 & 0.5042\\
        \cline{2-11}
        & (\linkps + \graphgs) $\otimes$ \abnorms $\otimes$ \ensembs & \bf 0.8350 & \bf 0.8750 & 0.6047 & \bf 0.8024 & \bf 0.9176 & \bf 0.9220 & \bf 0.7827 & \bf 0.6712 & \bf 0.5223 \\
        \hline
        \hline
        \multirow{3}{*}{Rule} & \bls & 0.5609 & 0.5499 & 0.4449 & 0.4752 & 0.5357 & 0.5585 & 0.2577 & 0.0741 & 0.5502 \\ 
        \cline{2-11}
        & \blbs & 0.9689 & 0.9666 & 0.9696 & 0.9389 & 0.9392 & 0.9071 & 0.9796 & 0.9766 & 0.9660 \\
        \cline{2-11}
        & (\linkps + \graphgs) $\otimes$ \abnorms $\otimes$ \ensembs & \bf 0.9845 & \bf 0.9885 & \bf 0.9946 & \bf 0.9865 & \bf 0.9841 & \bf 0.9832 & \bf 0.9934 & \bf 0.9796 & \bf 0.9724 \\
        \hline
    \end{tabular}
\end{table*}

\subsection{Detection Performance on Real-world Datasets}
We will first provide the overall performance comparison between our proposed pipeline EDoG and the state-of-the-art baselines~\cite{perozzi2016recommendation,rattigan2005case} with and without knowledge of the attacker's approach to demonstrate its efficacy, and then
present how each of our proposed detection approaches perform on different real-world datasets and attacks strategies. 

\textbf{Overall Performance without knowledge of attack type.}
Despite the efficacy of our approach compared with baselines, in many cases we do not know the attack approach.
Therefore, we need a uniform pipeline, \ensemb, to defend against the various types of attacks. The performance of \ensemb compared with the baselines is shown in Table \ref{tab:overall}. To avoid data dependency and make the comparison more clear, we average the results over datasets to evaluate the overall performance. As we can see, our approach EDoG outperforms the baselines on all the tasks. Also, we observe that defending against Single-edge attack and Multi-edges indirect attack are relatively easy tasks. Our pipeline can achieve over 0.8 AUC on these tasks. The other two attacks are relatively hard to defend. Some results of baseline approaches are even worse than random guess. As we discussed before, this may be due to the principle of the \emph{collective power of malicious edges}. Nevertheless, our approach can still get an over 0.7 detection AUC against such kind of attack since we use the scores of \abnorm to detect edges between high degree nodes. This again demonstrates the robustness of our general pipeline.

\textbf{Overall Performance with knowledge of attack type.}
As shown in Table \ref{tab:result-known}, our approach \ensembs outperforms the state-of-the-art baselines in most cases significantly when there is knowledge about the attack type.
The performance of different detection approaches matches with our intuition discussed above, and different models do well against different types of attacks. Hence, when we have knowledge of the attack approach, we can use the detection method that is good at dealing with it. This is also useful when we want to defend against specific type of attacks. In particular, \linkp + \graphg works well for Single-edge attack and Multi-edges direct attack on small-degree target nodes (degree $\leq 5$). For Multi-edges direct attack on large-degree target nodes, \abnorm is the appropriate choice. On Multi-edges indirect attack, it turns out that both \graphg and \linkp + \graphg would work well and we choose \linkp + \graphg to be coherent with that in Single-edge attack. For Meta attack, \ensemb is used since Meta-attack may contain various types of malicious edges. 
The average AUC of the proposed approaches is close to 0.85, demonstrating the efficacy of our detection pipeline \ensembs against different  attacks.

In order to explore the detection performance of different detection primitives, we show the results of each detection method that we proposed in Appendix~\ref{sec:app-diff-prim}. We will observe that graph generation-based method has superior performance for low-degree attacks while outlier detection method performs better on high-degree attacks. With our ensemble method \ensembs, the performance will be good for all tasks.

\begin{table*}[th]
    \centering
    \caption{The average AUC on synthetic dataset. BA stands for the scale-free graph generated by Barabasi-Albert algorithm. ER is the average result of the Erdos-Renyi graphs generated with $p_1=\frac{\ln n}n$ and $p_2=\frac{2\ln n}n$ respectively.}
    \vspace{-1mm}
    \label{tab:synth-add}
    \begin{tabular}{c|c|c|c|c|c|c|c|c}
        \hline
         & \multicolumn{2}{|c|}{Single-edge attack} & \multicolumn{2}{|c|}{Multi-edges direct attack} & \multicolumn{2}{|c|}{Multi-edges indirect attack} & \multicolumn{2}{|c}{Meta attack}\\
        \hline
        Dataset & BA & ER & BA & ER & BA & ER & BA & ER\\
        \hline
        \bls & 0.5000 & \bf 0.8465 & 0.5000 & 0.6146 & 0.7495 & 0.6535 & 0.5194 & 0.4550\\
        \hline
        \blbs & 0.6265 & 0.4298 & 0.2633 & 0.3077 & 0.1667 & 0.6100 & 0.6123 & 0.4322\\
        \hline
        \ensembs & \bf 0.9975 & 0.8289 & \bf 0.7896 & \bf 0.7397 & \bf 0.9437 & \bf 0.8247 & \bf 0.9626 & \bf 0.7985\\
        \hline
    \end{tabular}
    \vspace{-3mm}
\end{table*}

\subsection{Analysis on Synthetic Dataset}
Besides the real-world graph dataset, we also evaluate the attack approaches and our detection framework on the synthetic ER graphs and scale-free BA graph, aiming to obtain in-depth analysis of our detection approaches in the controlled environments. In this set of experiments, we perform in-depth analysis for how well the EDoG perform in terms of detecting adversarial edges, if the graph has certain properties to make the detection harder.
The result is shown in Table \ref{tab:synth-add}. As we can see, the performance of our detection approach over synthetic graphs is even better than that over real-world data. The performance on the scale-free graph, which is similar to real world citation network structure, reaches an average of above 92\% AUC on all tasks. The performance on Erdos-Renyi graphs are comparatively low, yet still achieve an average AUC of around 80\%. This shows that our approach does exploit the structure of scale-free graph to improve the performance. Also, we observe that our approach beats baselines on all Multi-edges direct attack and Meta attack, which we consider as hard tasks. For the other two tasks, the baseline can sometime outperform our approach.

Hence, we conclude that \ensembs can generalize to different graph structures. In addition, the baseline approaches can sometimes work well in clean synthetic dataset, but in complicated and noisy real-world graph structure, the robustness of our approach is superior.

\subsection{Visualization of Attack and Detection}
\begin{figure}[h]
    \centering
    \includegraphics[width=1.0\linewidth]{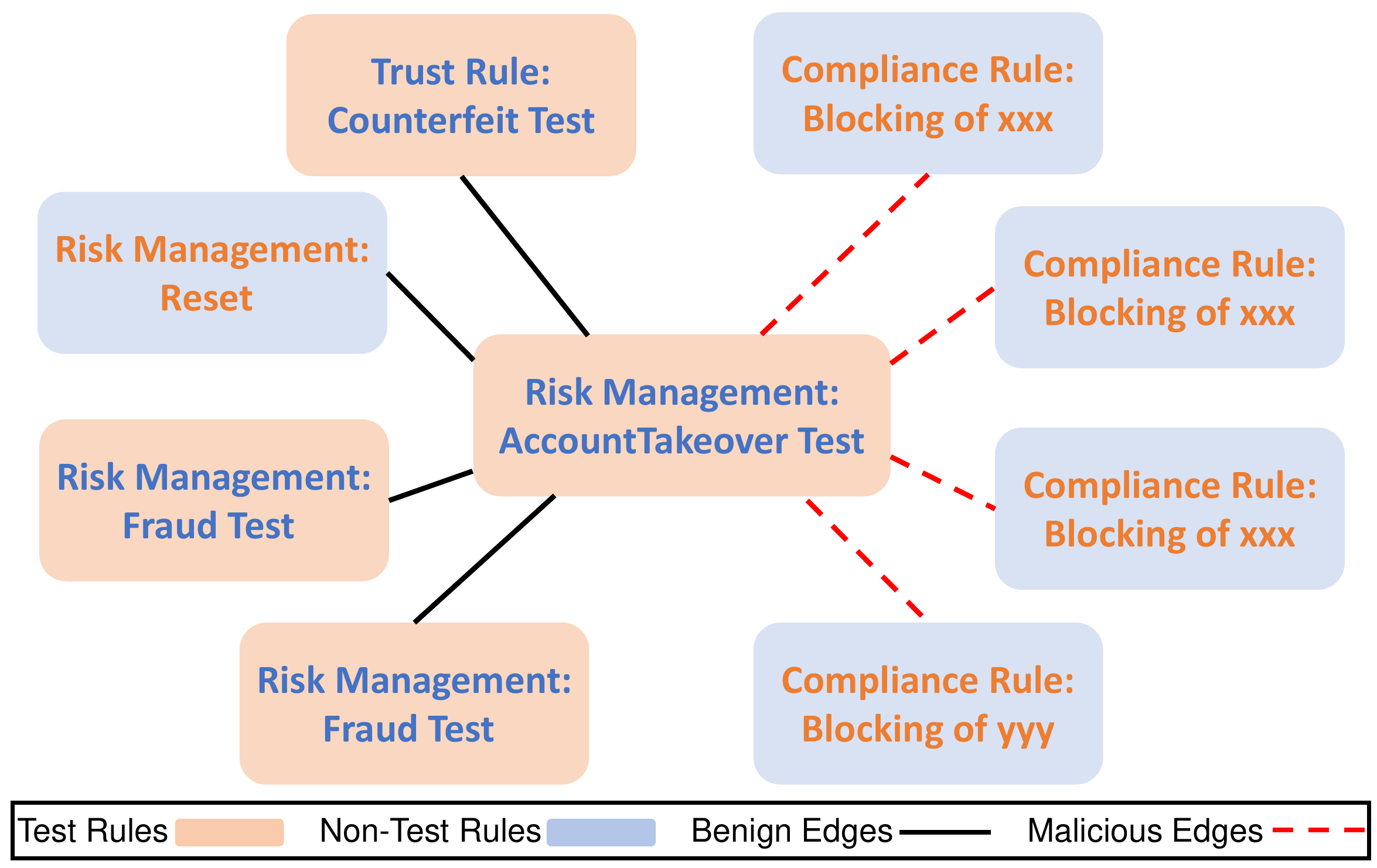}
    \caption{Visualization of one multi-edges direct attack on Rule dataset. The node in the middle is the target node. Our EDoG pipeline successfully detects all the injected malicious edges.}
    \label{fig:visual}
    \vspace{-5mm}
\end{figure}

To intuitively understand the adversarial attack and corresponding detection performance, we include a visualization of an attack and our detection on the Rule dataset as shown in Fig.~\ref{fig:visual}. The node in the middle is the target node and the attacker performed multi-edges direct attack to inject several malicious edges connected to it. We can observe that the target node is a test rule and the attacker's goal is to fool the model to predict it as non-test rule, so the malicious edges are all connected to the non-test rules. During detection, our EDoG pipeline detects such adversarial edges easily. All the malicious edges are detected with a high EDoG score. 

\subsection{Adaptive Attacks}

In this subsection, we consider the adaptive attack  where the attacker has full knowledge of our detection pipeline and will intentionally try to bypass our detection during their attack. We design the adaptive attack as follows: given the clean graph $G$, the attacker will first run the \ensembs pipeline and get the prediction scores for all the node pairs. Then, during the attack process, the attacker will only inject the malicious edges with a lower \ensembs prediction score (in practice we use only the lower 25\% edges). With such process, the attacker tries to only inject the edges that seem benign to the \ensembs system. Note that the edges are not guaranteed to evade the detection, since the \ensembs score will change after the injection of malicious edges.

We show the results of adaptive attacks as in Table~\ref{tab:adapt-atk} and compare with the standard attack which is allowed to inject arbitrary edges. In order to perform adaptive attack, the attacker's choice of malicious edges are restricted. Therefore, we can observe that the attack success rate is greatly reduced. In most cases the success rate is reduced by over 50\%. This shows that our detection pipeline is difficult to bypass even if the attacker has full knowledge of it.

\begin{table}[htbp]
    \centering
    \caption{Comparison of the attack success rate between standard attack and adaptive attack against \ensembs.}
    \label{tab:adapt-atk}
    \begin{tabular}{c|c|c|c|c}
         \toprule
         & Attack & Cora & Citeseer & Rule \\
         \hline
         & Single-edge & 0.475 & 0.463 & 0.246\\
         Standard & Multi-edges direct & 0.729 & 0.842 & 0.556\\
         Attack & Multi-edges indirect & 0.149 & 0.211 & 0.052\\
         & Meta & 0.376 & 0.333 & 0.240\\
         \hline
         \hline
         & Attack & Cora & Citeseer & Rule \\
         \hline
         & Single-edge & 0.070 & 0.217 & 0.146\\
         Adaptive & Multi-edges direct & 0.486 & 0.397 & 0.483 \\
         Attack & Multi-edges indirect & 0.087 & 0.066 & 0.033 \\
         & Meta & 0.026 & 0.022 & 0.000 \\

        
         \hline
    \end{tabular}
    \vspace{-3mm}
\end{table}

\subsection{Detecting Random Edges on Graphs}
In order to evaluate whether our pipeline is able to capture the behaviours of pure malicious attackers, we conduct another experiment where we add random edges into the graph. The detailed results and analysis are shown in Appendix \ref{sec:result-random}. We can observe that baseline approaches will consider random edges as some malicious behaviour, while ours will ignore those randomness and focus only on those malicious edges.


\section{Related Work}

\textbf{Adversarial attacks on graphs.}
Recently the robustness of machine learning models has been studied in numerous settings such as adversarial examples \cite{Szegedy2013Intriguing,xiao2018generating,xiao2018spatially,goodfellow2014explaining}.
Most of the researches on on a continuous input space (e.g. images and audios).
As for text domain, some of attacks are based on manually constructed perturbations (\cite{jia-liang:2017,Samanta2017Towards}) and \cite{Gong2018Adversarial} adopted the gradient attacking method to the embedding space of texts. 
As for graphs, performing attacks can be more difficult especially when we want to make perturbations on edges, since the gradient attacking method is not easy to be used here. 
So far there are several attacks proposed on graphs: some of the works attempted to attack the graph neural networks \cite{dai2018adversarial,zugner2018adversarial,zugner2018meta}, as \cite{dai2018adversarial} proposed three attack methods on both node classification and graph classification problems and \cite{zugner2018adversarial} developed an attack method targeting on a particular node based on greedy approximation scheme. Rather than reducing the classification result of a particular node,  \cite{zugner2018meta} attacked the overall performance of node classification tasks through meta learning method. Moreover, \cite{bojcheski2018adversarial,sun2018data}  studied the vulnerability of unsupervised node embedding models.

\textbf{Defense methods against adversarial attacks.}
Defense on neural networks is much harder compared with attacks (\cite{Meng2017MagNet}). Currently, most of the defense methods are based on (1) \textit{changing the architecture of deep learning models:} \cite{Papernot2016Distillation} leveraged distillation training techniques and reduce the magnitude of gradients between the pre-softmax
layer (logits) and softmax outputs. \cite{svoboda2018peernets} used the graph attention network (\cite{velickovic2018graph}) to harness information from a graph of peer samples to improve the robustness of network. (2) \textit{adversary training methods:}  \cite{goodfellow2014explaining} managed to augment the training dataset with adversarial examples. 
(3) \textit{data preprocessing:} \cite{Xu2017Feature} hardened the deep learning models against adversarial perturbations on images through reducing the color depth and smoothing on spatial domain. \cite{Meng2017MagNet,samangouei2018defensegan} attempted to filter the adversarial noise of the input samples with autoencoders and Generative Adversarial Networks (GAN) (\cite{NIPS2014_5423}) respectively. 
Although those approaches are somehow effective, they still have some limitations and the task of defending against adversarial attacks is still challenging. 

In this paper we focus on detecting malicious edges in a graph under attack. There are also works aiming to directly defend the attacks and mitigate the effect of malicious edges \cite{zhu2019robust}. Comparisong with defense approaches, our detection approach is superior in two properties: first, detecting the malicious edge can help identify the attacker. For example, in the case of social network we can find out which user is adding the malicious link. Second, the defense approaches are usually more prone to adversarial attack, since they incorporate the defense mechanism in the model and are therefore easier to be attacked as a whole when the attacker has knowledge of such mechanism.

\textbf{Robust Graph Neural Networks.}
Considering that the threat of adversarial attacks on GNNs is newly emereged, the research with regard to defense methods on graph neural networks is limited. \cite{zugner2019certifiable} proposes a certificate for GNN robustness under adversarial attack by changing node features. \cite{feng2019graph} proposes adversarial retraining pipeline to improve GNN model robustness and it also deals with node feature attack. The goal of these two works is different from ours since we are defending against graph structure attack. \cite{xu2019topology} also proposes an adversarial retraining pipeline to defend against graph structure attack. They evaluate over the meta attack and the retrained model still suffers from performance decrease. \cite{zhu2019robust} proposes a GNN structure where each layer is no longer deterministic but a stochastic transformation layer. They show that using this structure the model performance will increase under adversarial structure attack. However, when the number of added edges increases the model will still fail.

\textbf{Anomaly detection on graphs.}
There have been various researches on the topic of `\textit{Anomaly Detection}' or `\textit{Fraud Detection}' on graphs \cite{Akoglu2015Graph,jiang2014catchsync,shah2017many}. Note that the detection purpose is different from ours: anomaly detection on graphs aims to find nodes/edges that differ a lot from others. While the two proposed subtle adversarial attack on graphs appear to contain too small magnitude of perturbation to be detected, and here we focus on graph neural networks instead of general graph analysis. In addition, \cite{chen2018bootstrap} studied on how to fit a good model over graph data when only one graph is available. They focus on the classification performance while our work focuses on the detection of malicious edges, and we have completely no supervision. Our work is going to defend against graphical network models by detecting malicious edges with original approaches. \cite{deng2019batch,sun2019virtual} propose virtual adversarial training on GNNs as a way to improve model performance but they do not evaluate their model against adversarial attack.

\section{Conclusions}
Overall, we propose the first general detection pipeline \ensembs against the state-of-the-art attack on GNNs. We investigate into the attack and defense properties and find that different attack strategies led to different behaviors: malicious edges connecting low degree nodes will not likely to appear as outliers while the ones connecting high degree nodes will. 
Thorough experiments show that the average detection AUC of \ensembs can reach above 80\% and adaptive attacks are hard to be performed given the complexity of the detection pipeline. These results shed light on the design of robust GNNs against attacks on graph-structured data. 

\vspace{1mm}
\section*{Acknowledgements}
This work is partially supported by NSF grant No.1910100, NSF CNS No.2046726, a C3.ai DTI Award, and the Alfred P. Sloan Foundation.

\bibliographystyle{IEEEtranS}
\bibliography{ref}

\appendices

\newpage

\section{Benign accuracy}
\label{sec:app-benign-acc}
In Table~\ref{tab:benign-acc}, we show the benign accuracy of our GNN models on the different datasets.

\begin{table}[tbp]
    \centering
    \caption{The benign accuracy of the model on different datasets.}
    \label{tab:benign-acc}
    \begin{tabular}{c|c|c|c}
        \toprule
        Task & Cora & Citeseer & Rule \\
        \midrule
        Benign Accuracy & 0.819 & 0.697 & 0.859 \\
        \bottomrule
    \end{tabular}
\end{table}

\section{Traditional metrics}
\label{sec:app-other-metric}
In addition to the baselines mentioned in the paper, we also evaluate two heuristic metrics which are commonly used in link prediction tasks - the common neighborhood (CN) score and the Adamic-Adar (AA) score. The CN score will calculate the common neighbors between two nodes, while the AA score will calculate the inverse logarithmic degree centrality of the common neighbors. We show the detection performance of the metrics on Cora and Citeseer in Table~\ref{tab:other-metric}. We can observe from the tables that our approach can still outperform these baselines in detecting malicious edges.

\begin{table*}[htbp]
    \centering
    \caption{Detection performance comparison between \ensembs and traditional heuristic metrics against the malicious edges.}
    \label{tab:other-metric}
    \begin{tabular}{c|c|c|c|c|c}
        \toprule
        Task & Method & Single-edge attack & Multi-edge direct attack & Multi-edge indirect attack & Meta attack \\
        \midrule
        \multirow{3}{*}{Cora} & CN & 76.94 & 51.99 & 74.89 & 64.69 \\ 
        & AA & 76.94 & 48.55 & 74.61 & 64.59 \\
        & \ensembs & \bf 90.99 & \bf 84.91 & \bf 83.91 & \bf 68.84 \\
        \midrule
        \multirow{3}{*}{Citeseer} & CN & 31.29 & 46.05 & 65.40 & 51.43 \\ 
        & AA & 27.71 & 40.78 & 64.07 & 49.47 \\
        & \ensembs & 85.50 & 81.17 & 72.69 & 52.23 \\
        \bottomrule
    \end{tabular}
\end{table*}

\begin{table*}[htbp]
    \centering
    \caption{The AUC of the detection components we proposed against different types of attacks. Here we explicitly consider target node with various degrees to demonstrate the generalization of our detection methods. Different components have their own advantages under certain setting.}
    \vspace{-2mm}
    \label{tab:result-ours}
    \begin{tabular}{c|c|c|c|c|c|c|c|c|c|c}
        \toprule
        & & \multicolumn{2}{c|}{Single-edge attack} & \multicolumn{4}{c|}{Multi-edges direct attack} & \multicolumn{2}{c|}{Multi-edges indirect attack} & Meta attack \\
        \cline{2-11}
        & Node degree & $[1,5]$ & $(5,+\inf)$ & $[1,5]$ & $(5,10]$ & $(10,15]$ & $(15,+\infty)$ & $[1,10]$ & $(10,+\inf)$ & - \\
        \hline
        \multirow{5}{*}{Cora} & \linkps & 0.8960 & 0.8944 & 0.8319 & 0.6564 & 0.3787 & 0.3224 & 0.8506 & 0.8136 & 0.7935\\
        \cline{2-11}
        & \abnorms & 0.2946 & 0.5233 & 0.4326 & 0.7378 & 0.9144 & 0.9123 & 0.2122 & 0.3588 & 0.4379 \\
        \cline{2-11}
        & \graphgs & 0.6876 & 0.8535 & 0.6156 & 0.6864 & 0.4712 & 0.4643 & 0.6907 & 0.8156 & 0.5749\\
        \cline{2-11}
        & \linkps + \graphgs & 0.9088 & 0.9110 & 0.8319 & 0.6564 & 0.3787 & 0.3224 & 0.8368 & 0.8414 & 0.7568\\
        \cline{2-11}
        & \ensembs & 0.9478 & 0.8794 & 0.7618 & 0.6910 & 0.6368 & 0.6202 & 0.6985 & 0.8080 & 0.6884 \\
        \hline
        \hline
        \multirow{5}{*}{Citeseer} & \linkps & 0.8612 & 0.7612 & 0.5954 & 0.1633 & 0.2155 & 0.3181 & 0.6582 & 0.5924 & 0.3221 \\
        \cline{2-11}
        & \abnorms & 0.0858 & 0.5211 & 0.3457 & 0.8024 & 0.9176 & 0.9220 & 0.4498 & 0.3874 & 0.5939\\
        \cline{2-11}
        & \graphgs & 0.5547 & 0.5972 & 0.8137 & 0.6945 & 0.6834 & 0.5868 & 0.9068 & 0.7530 & 0.5232\\
        \cline{2-11}
        & \linkps + \graphgs & 0.8350 & 0.8750 & 0.6047 & 0.2505 & 0.4027 & 0.3910 & 0.7827 & 0.6712 & 0.3681 \\
        \cline{2-11}
        & \ensembs & 0.7051 & 0.6894 & 0.7219 & 0.5524 & 0.7217 & 0.6248 & 0.9018 & 0.6574 & 0.5223\\
        \hline
        \hline
        \multirow{5}{*}{Rule} & \linkps & 0.9845 & 0.9885 & 0.9370 & 0.9787 & 0.9631 & 0.9675 & 0.9932 & 0.9021 & 0.8658 \\
        \cline{2-11}
        & \abnorms & 0.9672 & 0.9836 & 0.9946 & 0.9865 & 0.9841 & 0.9832 & 0.9934 & 0.9503 & 0.9579 \\
        \cline{2-11}
        & \graphgs & 0.7869 & 0.8833 & 0.7979 & 0.7025 & 0.8175 & 0.7355 & 0.9032 & 0.8207 & 0.7880 \\
        \cline{2-11}
        & \linkps + \graphgs & 0.9481 & 0.9382 & 0.9331 & 0.8713 & 0.9015 & 0.8320 & 0.9119 & 0.9297 & 0.9315 \\
        \cline{2-11}
        & \ensembs & 0.9679 & 0.9767 & 0.9633 & 0.9297 & 0.9410 & 0.8971 & 0.9658 & 0.9413 & 0.9724 \\
        \hline
    \end{tabular}
    \vspace{-2mm}
\end{table*}

\section{Performance of Different Detection Primitives on Real-world Datasets}
\label{sec:app-diff-prim}
In this section we will explore in-depth on the detection performance against different types of attacks based on our proposed detection primitives.

\textbf{Single-edge attack.}
The result for Single-edge attack is shown in Table \ref{tab:result-ours}. As we can see, the link prediction-based approaches perform quite well. The best approach is \linkp+ \graphg, achieving an average of over 0.9 AUC over the tasks. 
When comparing the performance of the combination model \linkp+ \graphg with \graphgs, we see that the combination model significantly improves the detection performance as the \linkp filters out the malicious edges (i.e., AUC $>$ 0.5). This shows that our \graphg detection model can learn ``benign properties" from normal graphs and improve detection performance. On the other hand, a direct \graphg can achieve good performance in some cases. This shows that \graphg is a good yet unstable model, so a filtering algorithm can be combined to reduce the variation and improve its robustness.

In addition, We observe that the performance of \linkps+ \graphgs in Cora is slightly better than in Citeseer while Rule has the best performance. This is because that the graph of Citeseer is more sparse than Cora (with more nodes and fewer edges), and Rule is the most densne graph. After filtering, the information contained in Citeseer dataset is reduced and therefore it is harder for the model to learn useful patterns.

\textbf{Multi-edge direct attack.}
The result for Multi-edge direct attack is shown in Table \ref{tab:result-ours}. We see that as the node degree increases, approaches related to link prediction algorithms perform no better than random guessing. As we have discussed in Section \ref{sec:intuition-od}, we attribute this to a principle of the  `collective power of malicious edges'. By contrast, this collective power does not fool the \abnorm approach; its average AUC is over 0.85 when target node degree is larger than five, and 0.9 when it is larger than ten.

\textbf{Multi-edge indirect attack.}
The result for Multi-edge indirect attack is shown in Table \ref{tab:result-ours}. We see that the detection AUC decreases compared with Single-edge attack. This is reasonable: since more malicious edges are added, the defense would be more challenging. We observe that the \linkp approach is affected the most. Therefore, the filtering algorithm does not help to improve the performance of \graphg. The best approach here is \graphg. One surprising observation is that the performance in Citeseer is better than Cora. We think that this phenomenon is also because of the sparsity: malicious edges tend to accumulate in a small neighbourhood of the target node, and the sparsity induces that subgraphs of nodes far away from the target node will not contain malicious edges. Therefore, the proportion of benign subgraphs will increase and therefore the trained generative detection model can learn a better pattern of benign properties.

\textbf{Meta attack.}
The result for Meta attack is shown in Table \ref{tab:result-ours}. We see that none of the approaches shows a very good performance against such kind of attack, especially over the Citeseer dataset. This is because this meta attack may be a combination of different type of attacks, and a large number of malicious edges are added (5\%). Nevertheless, we can still see that our \ensemb pipeline reaches acceptable performance on over the tasks. The detection performance on the Rule dataset is still very good, since it is a dense graph and contains more information for detection.

\section{Results on Detecting Random Edges on Graphs}
\label{sec:result-random}
\begin{table*}[htbp]
    \centering
    \caption{The ratio of detecting non-random edges on Cora and Citeseer dataset. Larger value indicates that the model does not detect the added random edges, and 50\% means the detection method will not distinguish random and benign edges, which is desired (Results that outperform the baselines by more than 3\% are highlighted).}
    \label{tab:result-random}
    \begin{tabular}{c|c|c|c|c|c|c|c|c|c|c}
        \hline
        & \multicolumn{5}{c|}{Cora} & \multicolumn{5}{c}{Citeseer} \\
        \hline
        \#Random edge & 1 & 2 & 4 & 8 & 16 & 1 & 2 & 4 & 8 & 16 \\
        \hline
        \bls & 10.98\% & 17.54\% & 26.15\% & 11.88\% & 15.00\% & 4.35\% & 27.42\% & 11.37\% & \bf 30.51\% & 29.11\% \\
        \hline
        \blbs & 2.16\% & 7.74\% & 19.83\% & 7.18\% & 5.47\% & 7.07\% & 4.16\% & 7.11\% & 13.61\% & 8.56\% \\
        \hline
        \ensembs & 9.64\% & \bf 52.35\% & \bf 34.52\% & \bf 30.74\% & 17.43\% & \bf 36.32\% & \bf 42.21\% & \bf 23.32\% & 22.21\% & \bf 45.19\% \\
        \hline
    \end{tabular}
\end{table*}

In this experiment, we randomly choose 1, 2, 4, 8, and 16 unconnected node pairs in Cora and Citeseer datasets to add random edges. We then run our detection pipeline as well as the baseline approaches on the graph to see whether these random edges will be identified as the malicious edges.
In Table \ref{tab:result-random}, we show the ratio of detecting non-random edges. Hence, larger value means that the detection model will not be distracted by the added random edges. In particular, 50\% means that the detection method will not distinguish random and benign edges which is a desired property.

We can observe that baseline approaches tend to detect random edges as malicious edges while EDoG will ignore most random added edges.
This is because these baseline approaches are essentially designed to detect abnormality in the graphs and therefore they will recognize the abnormal behaviour of the randomly added edges. 
On the other hand, our detection pipeline \ensemb will not identify the random edges as malicious ones. In some cases the value is near 50\% which means that the model views random edge to be similar as normal ones. This is ideal for our goal since our model will not be distracted by the random edges in the task of malicious edge detection. Comparing the result with the baseline approaches, we claim that our \ensemb pipeline can not only detect malicious behaviors of the attacker, but also resilient against random added edges, which has great potential to improve the robustness of GNNs.



\end{document}